\documentclass[10pt,conference]{IEEEtran} 
\usepackage[T1]{fontenc} 
\usepackage{cite} 
\usepackage{amsmath,amssymb,amsfonts} 
\usepackage{graphicx} 
\usepackage{textcomp} 
\usepackage{xcolor} 
\usepackage{booktabs} 
\usepackage{array} 
\usepackage{url} 
\usepackage{microtype} 
\usepackage{capt-of} 
\usepackage{cuted} 
\usepackage{stfloats} 
\usepackage{newtxtext,newtxmath}
\definecolor{linkteal}{rgb}{0.15,0.45,0.55}
\usepackage[colorlinks=true,
            urlcolor=linkteal,
            linkcolor=black,
            citecolor=black]{hyperref}

\IEEEoverridecommandlockouts 
\makeatletter
\newcommand\blfootnote[1]{\begingroup\renewcommand\thefootnote{}\footnote{#1}\addtocounter{footnote}{-1}\endgroup}
\makeatother
\graphicspath{{figures/}} 
\newcommand{\method}{SymQNet} 
\newcommand{\mse}{\mathrm{MSE}} 
\newcommand{\E}{\mathbb{E}} 
\newcommand{\desctext}{\normalsize} 

\begin{document} 
\raggedbottom

\title{\method{}: Amortized Acquisition for Low-Latency Adaptive Hamiltonian Learning} 

\author{
  \IEEEauthorblockN{Yash Vardhan Tomar}
  \IEEEauthorblockA{\textit{Purdue University}\\ West Lafayette, United States\\ tomar4@purdue.edu}
  \and
  \IEEEauthorblockN{Dheeraj Peddireddy}
  \IEEEauthorblockA{\textit{Purdue University}\\ West Lafayette, United States\\ dpeddire@purdue.edu}
}
\maketitle

\blfootnote{\copyright~2026 IEEE. Personal use of this material is permitted. Permission from IEEE must be obtained for all other uses, in any current or future media, including reprinting/republishing this material for advertising or promotional purposes, creating new collective works, for resale or redistribution to servers or lists, or reuse of any copyrighted component of this work in other works.}

\begin{abstract}
Adaptive Hamiltonian learning is central to calibrating and characterizing quantum devices. In an adaptive controller, choosing the next experiment is itself a computation. Bayesian design rules are recomputed after every posterior update, and that step can take seconds. Across hundreds of shots, those seconds become a significant wall-clock cost for adaptivity. We introduce \method{}, an amortized reinforcement-learning approach for low-latency adaptive Hamiltonian learning. \method{} learns a posterior-conditioned acquisition policy offline, then uses a fast policy forward pass online while retaining Bayesian posterior feedback. On transverse-field Ising benchmarks, \method{} substantially reduces acquisition latency relative to bounded Fisher-information search and bounded two-step Bayesian active learning by disagreement (BALD). At five qubits, it reduces acquisition-only decision latency by $47.1\times$ and $72.6\times$ relative to these online baselines; at twelve qubits, full simulated steps take $1.02$ s for \method{} versus $13.27$ s for bounded two-step BALD. Overall, we show that learned acquisition can make adaptive Hamiltonian learning practical for repeated low-latency workloads.
\end{abstract}

\par\vspace{4pt}


\section{Introduction}

Quantum computers are compelling because they can, in favorable settings, represent and manipulate physical states beyond the reach of classical hardware. That promise depends on devices whose behavior is known well enough to control. Calibration and sensing therefore become hardware-level prerequisites. Before a quantum algorithm can be trusted, the couplings and fields of the processor have to be learned from data, and Hamiltonian learning estimates those quantities from measurements. Adaptive experiment design is attractive here because each new measurement can use the current posterior, so later shots can probe uncertainties exposed by earlier outcomes \cite{wiebe2014hamiltonian,granade2012robust,huszar2012adaptive}. This makes Hamiltonian learning a sequential decision problem in which the learner acts, observes, updates the belief, and chooses again.

Classical Bayesian experimental design provides several acquisition rules for this loop. Fisher-information search and Bayesian active learning by disagreement (BALD) \cite{houlsby2011bald} score candidate measurements using the posterior. Sequential Monte Carlo (SMC) supplies an approximate Bayesian update for nonlinear quantum models. The computational burden comes from repeated posterior-conditioned scoring. After each posterior update, the acquisition rule must score another set of candidate experiments. For $N$ qubits, that score depends on particles, measurement bases, evolution times, and any lookahead tree. Even a capped online search can become the dominant runtime component of the controller.

We train a policy that amortizes acquisition across related Hamiltonian-learning instances. This view follows Deep Adaptive Design and reinforcement-learning (RL) formulations of sequential Bayesian experimental design \cite{foster2021deep,blau2022optimizing}. \method{} applies it to transverse-field Ising model (TFIM) Hamiltonian learning. The policy reads an SMC belief state, graph context, and measurement history, then selects the next qubit, basis, and evolution time. Training uses proximal policy optimization (PPO) \cite{schulman2017ppo}.

\begin{figure}[!ht]
\centering
\includegraphics[width=\columnwidth]{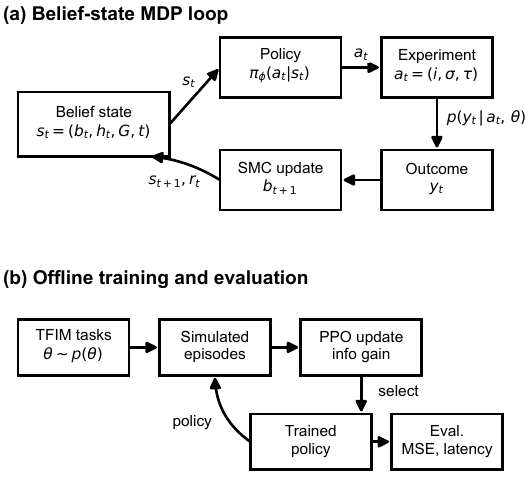}
\refstepcounter{figure}\label{fig:mdp}
\par\vspace{4pt}
\noindent\parbox{\columnwidth}{\raggedright\desctext\textbf{Fig. \thefigure.} Belief-state Markov decision process (MDP) and method flow. Here $s_t=(b_t,h_t,G_N,t)$ is the state (posterior, history, graph, time); $\theta\sim p(\theta)$ is the task prior, $\pi_\phi$ the policy, $a_t=(i,\sigma,\tau)$ a qubit/basis/time choice, $y_t$ an outcome, and $r_t$ the reward. In the text, $G_N$ denotes the $N$-qubit chain graph.}
\end{figure}

Our experiments report acquisition latency separately from final estimation accuracy. Fixed schedules and a Deep Adaptive Design (DAD)-style transformer serve as the main accuracy references in the controlled matched-prior benchmark. Against bounded online acquisition, \method{} maintains competitive parameter mean-squared error (MSE) and reduces decision time by orders of magnitude on eight-, ten-, and twelve-qubit TFIM instances. Figure~\ref{fig:mdp} shows the belief-state loop and offline training flow used throughout the paper.

\section{Preliminaries}

\subsection{Hamiltonian-Learning MDP}

The task distribution isolates adaptive measurement selection in a controlled many-body model with a discrete action space that grows with $N$. Each episode uses an $N$-qubit one-dimensional TFIM with Pauli $X_i,Z_i$, couplings $J_i$, fields $h_i$, and parameter vector $\theta$ \cite{pfeuty1970ising},
\begin{equation}
H(\theta) = \sum_{i=1}^{N-1} J_i Z_i Z_{i+1} + \sum_{i=1}^{N} h_i X_i,
\end{equation}
where $\theta=(J_1,\ldots,J_{N-1},h_1,\ldots,h_N)$ has dimension $d=2N-1$ with entries sampled independently from $U(0.5,1.5)$. An action chooses a qubit, a single-qubit Pauli basis, and one of $m=5$ evolution times $\tau\in\mathcal{T}=\{\tau^{(1)},\ldots,\tau^{(m)}\}$, equally spaced in $[0.1,1.0]$ in natural units ($\hbar=1$), giving $3Nm$ actions. The prior range, five time choices, and horizon $T=36$ give a positive-coupling TFIM with a linearly growing action set and repeated posterior updates. After an action, the simulator samples a measurement outcome with a fixed shot budget, and a sequential Monte Carlo (SMC) posterior with $P=256$ particles approximates the Bayesian belief update, using systematic resampling when effective sample size falls below $0.6P$ to limit weight degeneracy \cite{doucet2001sequential}.

Because the true parameter vector is latent, the posterior is used as a belief state, giving a finite-horizon belief-state MDP. Let $b_t(\theta)$ be the SMC posterior, $h_t=\{(a_\ell,y_\ell)\}_{\ell<t}$ the experiment history, and $G_N$ the $N$-node chain graph whose nodes are the qubits (each carrying per-qubit belief and observation features) and whose edges are the nearest-neighbor bonds $(i,i{+}1)$. The state and action are
\begin{equation}
s_t=(b_t,h_t,G_N,t),\qquad
a_t=(i_t,\sigma_t,\tau_t)\in\mathcal{A}_N ,
\end{equation}
where $\mathcal{A}_N$ is the action set for $N$ qubits, $i_t$ is the measured qubit, $\sigma_t\in\{X,Y,Z\}$ is the basis, and $\tau_t\in\mathcal{T}$ is one of the $m$ evolution times. The transition uses the measurement likelihood $p(y\mid \theta,a_t)$ in the Bayesian filtering step
\begin{equation}
y_t\sim p(y\mid \theta,a_t),\qquad
b_{t+1}(\theta)\propto p(y_t\mid \theta,a_t)b_t(\theta).
\end{equation}
The action $a_t=(i_t,\sigma_t,\tau_t)$ evolves the fixed initial state $\rho_0=\lvert0\rangle^{\otimes N}$ to $\lvert\psi_\theta\rangle=U_\theta(\tau_t)\lvert0\rangle^{\otimes N}$ with $U_\theta(\tau)=e^{-iH(\theta)\tau}$, and measures qubit $i_t$ with the single-qubit Pauli operator $\sigma_{i_t}\in\{X_{i_t},Y_{i_t},Z_{i_t}\}$. A single shot returns a $\pm1$ outcome with $\Pr(+1)=\tfrac12\big(1+(1-2p)\,\langle\psi_\theta|\sigma_{i_t}|\psi_\theta\rangle\big)$ under symmetric readout-flip rate $p$, and the observation $y_t\in[-1,1]$ is the empirical mean of these $\pm1$ outcomes over the shot budget.
The SMC posterior $b_t$ is represented by $P=256$ weighted particles $\{(\theta^{(p)},w_t^{(p)})\}_{p=1}^{P}$ with weighted mean $\mu_t$ and covariance $\Sigma_t$. During training, the policy $\pi_\phi(a_t\mid s_t)$ with parameters $\phi$ is optimized with an information-gain reward and discount $\gamma=0.99$, where the expectation $\E_{\pi_\phi}$ is taken over the task prior $\theta\sim p(\theta)$, the policy-induced action sequence, and the stochastic measurement outcomes and SMC resampling. The per-step reward is the nonnegative reduction in the Gaussian (moment-matched) differential entropy $\mathcal{H}(\Sigma)=\tfrac12\big(d\,[1+\log 2\pi]+\log\det\Sigma\big)$ of the posterior, and PPO maximizes the discounted objective $J(\phi)$,
\begin{equation}
r_t=\max\!\big(0,\;\mathcal{H}(\Sigma_t)-\mathcal{H}(\Sigma_{t+1})\big),\qquad
J(\phi)=\E_{\pi_\phi}\Big[\sum_{t=0}^{T-1}\gamma^t r_t\Big].
\end{equation}
This entropy reduction is a moment-matched estimate of the posterior information gain $D_{\mathrm{KL}}(b_{t+1}\|b_t)$; we compute it from the particle covariance rather than as an exact particle KL to keep the reward stable under resampling. Information gain gives a dense, posterior-local reward because every SMC update supplies signal without revealing the simulator's latent $\theta$. A direct MSE reward is available only at the end of simulated episodes and depends on the true parameter, which the controller never observes in deployment; we therefore train on this information-gain surrogate, reserve final MSE for held-out evaluation, and report the surrogate gap explicitly.

Held-out accuracy is reported as final parameter MSE,
\begin{equation}
\mse(\hat{\theta}_T,\theta)=\|\hat{\theta}_T-\theta\|_2^2/d,
\end{equation}
where $\hat{\theta}_T=\mu_T$ is the mean of the final SMC posterior $b_T$ and $d$ is the number of unknown parameters. The reported decision latency starts after the shared simulator and SMC state are available and stops after the acquisition rule chooses an action. Likelihood simulation, measurement time, and particle updates sit outside that timer for every method; total simulated step time is reported where it affects interpretation.
Within each benchmark, methods are evaluated on matched held-out task instances so MSE ratios and latency comparisons are paired across policies.

\section{\method{} and Acquisition Baselines}

\subsection{\method{} Policy}

\method{} is a learned adaptive measurement policy for the loop in Fig.~\ref{fig:mdp}. Its input matches the belief-state MDP variables, with observation history plus the chain graph and SMC posterior summaries. A variational autoencoder (VAE) compresses measurement-history features. Graph layers encode the qubit chain, and a transformer summarizes temporal history. The policy outputs logits over the discrete action set. The MDP form makes RL a direct tool here: each rollout follows the experimental loop, observing a belief state, choosing a measurement, receiving information gain, and continuing. PPO handles this loop without differentiating the simulator or SMC update. The five-qubit main and ablation policies use 2500 PPO updates with 64 rollout steps per update; scaling policies use 300 updates under the same rollout horizon so larger-system training remains computationally feasible. Held-out validation episodes select the learned policy, and the five-qubit benchmark uses five training seeds. Final evaluation uses parameter MSE, so information gain is treated as a surrogate reward and checked empirically.

At deployment, SMC still performs the posterior update, but the acquisition step is a policy forward pass. This preserves Bayesian filtering while replacing online acquisition optimization with amortized inference. The DAD-style neural comparator uses the same PPO and validation budget but removes the graph encoder, variational embedding, and SMC feedback; it observes the experiment history through a transformer-style sequential design model \cite{foster2021deep}.

\begin{table*}[!t]
\centering
\normalsize
\refstepcounter{table}\label{tab:main}
\begingroup
\setlength{\tabcolsep}{3.2pt}
\renewcommand{\arraystretch}{1.0}
\begin{tabular}{llrlll}
\toprule
Shots & Baseline & Episodes & MSE ratio & Speedup & $p_{\mathrm{BH}}$ \\
\midrule
128 & DAD-style Transformer & 2500 & 1.04 [1.03, 1.05] & 0.555 [0.554, 0.556] & $2.2\times10^{-18}$ \\
128 & Optimized fixed & 2500 & 2.38 [2.31, 2.47] & 0.0105 [0.00341, 0.0193] & $<10^{-300}$ \\
128 & Fisher-info (bounded) & 2500 & 1.97 [1.91, 2.03] & \textbf{72.6 [72.4, 72.8]} & $<10^{-300}$ \\
128 & 2-step BALD (bounded) & 2500 & 1.32 [1.29, 1.36] & \textbf{47.1 [47, 47.3]} & $2.9\times10^{-116}$ \\
512 & DAD-style Transformer & 2500 & 1.01 [1, 1.02] & 0.544 [0.543, 0.545] & 0.12 \\
512 & Optimized fixed & 2500 & 2.79 [2.62, 2.97] & 0.01 [0.0032, 0.0185] & $<10^{-300}$ \\
512 & Fisher-info (bounded) & 2500 & 2.29 [2.2, 2.38] & \textbf{83.7 [83.4, 84]} & $<10^{-300}$ \\
512 & 2-step BALD (bounded) & 2500 & 1.42 [1.38, 1.46] & \textbf{53.7 [53.5, 53.9]} & $5.7\times10^{-148}$ \\
\bottomrule
\end{tabular}

\endgroup
\par\vspace{4pt}
\noindent\parbox{0.88\textwidth}{\raggedright\desctext\textbf{Table \Roman{table}.} Five-qubit paired benchmark. MSE ratio is \method{} over baseline; speedup is baseline decision latency over \method{} decision latency; $p_{\mathrm{BH}}$ is the Benjamini--Hochberg-adjusted paired Wilcoxon p-value.}
\par\vspace{2pt}
\end{table*}

\subsection{Baselines and Bounded Online Acquisition}

The baselines separate fixed scheduling, learned amortized design, and online acquisition. Random measurement selection tests the value of policy structure. Cyclic and optimized fixed schedules test whether adaptivity is needed for the task distribution; the optimized fixed schedule is computed once from a prior-center Fisher-greedy search. The DAD-style transformer tests a generic neural sequential-design policy after removing graph, variational, and posterior-summary inputs.

Bounded Fisher-information search and bounded two-step BALD test the online-acquisition alternative. Exact online versions are computationally impractical at the scaling sizes considered here, so the bounded variants are reported explicitly. The bounded Fisher-information rule scores four candidate actions per decision with a particle-linearized Fisher objective. The bounded two-step BALD rule scores up to three present candidates; after retaining two lookahead candidates, it uses one predictive observation sample. These caps make paired comparisons tractable. The comparison treats Fisher and BALD as bounded online-acquisition references with stated budgets.

Each comparison fixes the simulator, task sequence, SMC update, and shot budget; only the acquisition rule changes. Unless otherwise noted, measurements include symmetric 2\% readout-flip noise, a mild readout-error model that avoids a noiseless-only claim. Richer relaxation, dephasing, and crosstalk channels enter through the simulator likelihood and SMC update, and remain future hardware-facing stress tests. Five-qubit comparisons use exact statevector simulation with a 32--512 shot sweep; 128 shots is the scaling point, and each seed uses 500 held-out episodes. Scaling uses 100 episodes per qubit count and a matrix-product-state backend \cite{vidal2004tebd}. All timings use CPU-only execution on x86-64 cluster nodes.

\section{Evaluating and Understanding Acquisition Cost}

\begin{figure*}[t]
\centering
\includegraphics[width=0.95\textwidth]{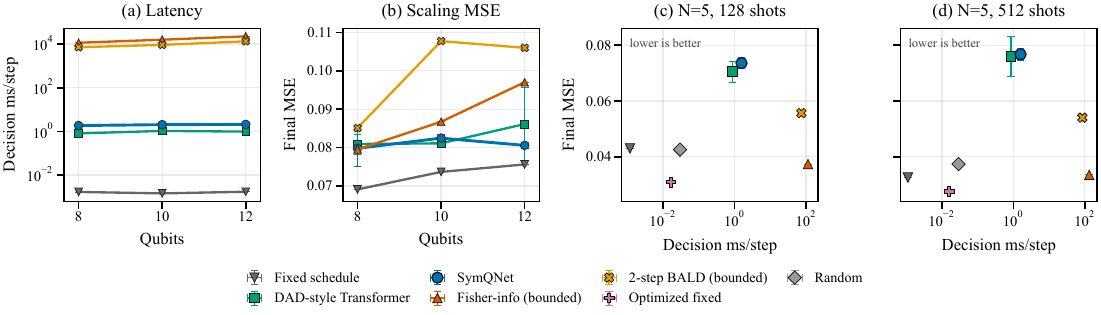}
\refstepcounter{figure}\label{fig:main}
\par\vspace{4pt}
\noindent\parbox{0.97\textwidth}{\raggedright\desctext\textbf{Fig. \thefigure.} Main evidence across scales. (a,b) On 8-, 10-, and 12-qubit tasks, \method{} remains near millisecond-scale decisions; bounded Fisher-information search and bounded two-step BALD take seconds to tens of seconds. \method{} and DAD-style policies use three seeds. (c,d) Five-qubit Pareto views show low \method{} decision cost, with fixed and neural baselines reaching lower final MSE.}
\end{figure*}

\subsection{Small-$N$ Tests Reveal a Speed--Accuracy Tradeoff}

The five-qubit benchmark is the primary small-$N$ diagnostic. Figure~\ref{fig:main}(c,d) gives the Pareto view, and Table~\ref{tab:main} reports paired comparisons. In Table~\ref{tab:main}, MSE ratios above one favor the baseline; speedups above one indicate slower baseline decisions. At 128 shots, bounded two-step BALD and bounded Fisher-information search give MSE-ratio/speedup pairs of 1.32/47.1 and 1.97/72.6.

The optimized fixed schedule is the most accurate small-$N$ reference on this statevector task, and the DAD-style comparator is the demanding neural baseline. At 512 shots, DAD is statistically indistinguishable from \method{} in MSE and has lower decision latency. \method{} reduces the decision cost of posterior-dependent acquisition, while fixed and neural baselines set the accuracy boundary in this matched-prior study. We expect the graph and SMC inputs to matter most when the geometry or prior shifts, or when the chain assumption is relaxed.

\subsection{Scaling Exposes the Online-Acquisition Bottleneck}

The scaling benchmark tests whether latency remains flat as action spaces grow on $N=8,10,12$ TFIM tasks with five evolution-time choices and 128 shots. \method{} and the DAD-style transformer are trained with three seeds.

\par\vspace{2pt}
\begin{center}
\begin{minipage}{0.92\columnwidth}
\centering
\normalsize
\refstepcounter{table}\label{tab:claim}
\begingroup
\renewcommand{\arraystretch}{1.04}
\begin{tabular*}{\linewidth}{@{}l@{\extracolsep{\fill}}l@{}}
\toprule
Quantity & Value \\
\midrule
Reference baseline & 2-step BALD (bounded) \\
SymQNet latency slope & 0.300 \\
Baseline latency slope & 1.519 \\
Latency slope ratio & \textbf{0.198} \\
Worst MSE ratio & \textbf{0.938} \\
\bottomrule
\end{tabular*}

\endgroup
\par\vspace{4pt}
\noindent\parbox{\linewidth}{\raggedright\desctext\textbf{Table \Roman{table}.} Scaling summary against bounded two-step BALD. MSE ratio is \method{} over baseline.}
\end{minipage}
\end{center}
\par\vspace{8pt}

\par\smallskip
At twelve qubits, \method{} takes 2.09 ms/decision, compared with 13.27 s for bounded two-step BALD and 22.32 s for bounded Fisher-information search, reductions of roughly $6.4\times 10^3$ and $1.1\times 10^4$. Table~\ref{tab:claim} summarizes the scaling result.

Relative to bounded two-step BALD, the decision-latency log-slope ratio is 0.20 and the worst MSE ratio is 0.94, where the decision-latency log-slope is the slope of a least-squares fit of $\log$ mean decision latency on $\log N$ over $N\in\{8,10,12\}$ and the ratio is \method{} over the baseline. Full simulated step time at twelve qubits is 1.02 s/step for \method{}, 13.27 s/step for bounded two-step BALD, and 22.32 s/step for bounded Fisher-information search. Figure~\ref{fig:main}(b) shows fixed schedules lowest-MSE and the DAD-style transformer competitive. Scaling bounded-acquisition points are single-run estimates, so the scaling claim is limited to flatter decision-latency growth.

\subsection{Architecture Checks Show Compression Headroom}

We conducted architecture ablations to compare the added representation against simpler policy classes. Table~\ref{tab:ablation} reports the main-shot-budget results.

\par\smallskip
\begin{center}
\begin{minipage}{0.96\columnwidth}
\centering
\normalsize
\refstepcounter{table}\label{tab:ablation}
\begingroup
\setlength{\tabcolsep}{4pt}
\renewcommand{\arraystretch}{1.05}
\begin{tabular}{lccc}
\toprule
Variant & Full MSE & Variant MSE & Delta \\
\midrule
No VAE & 0.0718 & 0.0726 & +1.1\% \\
No graph encoder & 0.0718 & 0.0740 & \textbf{+3.0\%} \\
MLP, no transformer & 0.0718 & 0.0709 & -1.3\% \\
\bottomrule
\end{tabular}

\endgroup
\par\vspace{4pt}
\noindent\parbox{\linewidth}{\raggedright\desctext\textbf{Table \Roman{table}.} Architecture ablations at 128 shots. Deltas are relative to the full policy; MLP means multilayer perceptron.}
\end{minipage}
\end{center}
\smallskip

At 128 shots, removing the graph encoder increases MSE by 3.0 percent and removing the VAE increases it by 1.1 percent; the compact MLP row removes the transformer and improves by 1.3 percent. The architecture signal is limited on this benchmark, which keeps the result centered on acquisition cost rather than one specific encoder stack. Smaller policies also cut forward-pass floating-point operations (FLOPs), giving a path toward microsecond-scale control. A reward diagnostic found weak alignment between cumulative information gain and final MSE (Pearson $-0.061$ to $0.069$, Spearman $-0.124$ to $0.006$ over 32--512 shots), which points to MSE-aligned rewards as the next training target.

\section{Conclusion and Future Work}

We introduced \method{}, a belief-state reinforcement-learning policy that keeps SMC posterior updates in the loop while replacing per-state Fisher/BALD scoring with a learned action rule. On TFIM benchmarks, this shift reduced acquisition latency by orders of magnitude relative to bounded online acquisition; at twelve qubits, full simulated steps were faster than bounded two-step BALD.

\method{} adds posterior-aware acquisition latency under paired accuracy checks, with optimized fixed schedules and the DAD-style transformer setting the accuracy boundary. Future work should replicate bounded scaling, add hardware timings, compress the policy, and train with rewards tied to final MSE.

\noindent\textbf{Data and code availability statement.} The depicted data and the used code can be found at \url{https://github.com/YTomar79/symqnet_quantum}.

\makeatletter
\let\oldthebibliography\thebibliography
\let\endoldthebibliography\endthebibliography
\renewenvironment{thebibliography}[1]{%
  \oldthebibliography{#1}%
  \fontsize{7.6pt}{7.65pt}\selectfont
  \setlength{\itemsep}{0pt}%
  \setlength{\parsep}{0pt}%
  \setlength{\parskip}{0pt}%
}{\endoldthebibliography}
\makeatother
\begingroup
\bibliographystyle{IEEEtran}
\bibliography{references}
\endgroup

\end{document}